\documentclass[journal]{IEEEtran}

\ifCLASSOPTIONcompsoc
  \usepackage[nocompress]{cite}
\else
  \usepackage{cite}
\fi

\ifCLASSINFOpdf
\else
\fi

\usepackage{graphics} 
\graphicspath{{figure/}}
\usepackage{epstopdf}

\usepackage{epsfig} 
\usepackage{mathptmx} 
\usepackage{times} 
\usepackage{amsmath} 
\usepackage{amssymb}  
\usepackage{subfigure}
\usepackage{amsthm}
\usepackage{algorithm}
\usepackage{algpseudocode}
\usepackage{enumerate}
\usepackage{colortbl}
\usepackage{multirow}
\usepackage{rotating}

\theoremstyle{plain}
\newtheorem{theorem}{Theorem}

\theoremstyle{definition}
\newtheorem{definition}{Definition}

\theoremstyle{definition}

\newtheorem{assumption}{Assumption}

\newtheorem{remark}{Remark}
\newtheorem{lemma}{Lemma}
\newtheorem{corollary}{corollary}

\begin{document}

\title{Data-Driven Inverse Reinforcement Learning for Expert-Learner Zero-Sum Games}

\author{\vskip 1em{
		Wenqian Xue,
        Bosen Lian,
		Jialu Fan,
        Tianyou Chai,
        and Frank L. Lewis
        }
\IEEEcompsocitemizethanks{\IEEEcompsocthanksitem

}


\thanks{Wenqian Xue, Jialu Fan, Tianyou Chai are with the State Key Laboratory of Synthetical Automation for Process Industries and International Joint Research Laboratory of Integrated Automation, Northeastern University, Shenyang 110819, China. (e-mail: xuewenqian23@163.com, fanjialu@gmail.com, tychai@mail.neu.edu.cn).}

\thanks{Bosen Lian and Frank L. Lewis are with the UTA Research Institute, the University of Texas at Arlington, Texas 76118, USA. (email: bosen.lian@mavs.uta.edu; lewis@uta.edu).}

}

\IEEEtitleabstractindextext{%
\begin{abstract}
In this paper, we formulate inverse reinforcement learning (IRL) as an expert-learner interaction whereby the optimal performance intent of an expert or target agent is unknown to a learner agent. The learner observes the states and controls of the expert and hence seeks to reconstruct the expert's cost function intent and thus mimics the expert's optimal response. Next, we add non-cooperative disturbances that seek to disrupt the learning and stability of the learner agent. This leads to the formulation of a new interaction we call zero-sum game IRL. We develop a framework to solve the zero-sum game IRL problem that is a modified extension of RL policy iteration (PI) to allow unknown expert performance intentions to be computed and non-cooperative disturbances to be rejected. The framework has two parts: a value function and control action update based on an extension of PI, and a cost function update based on standard inverse optimal control. Then, we eventually develop an off-policy IRL algorithm that does not require knowledge of the expert and learner agent dynamics and performs single-loop learning. Rigorous proofs and analyses are given. Finally, simulation experiments are presented to show the effectiveness of the new approach.
\end{abstract}

}

\maketitle

\IEEEdisplaynontitleabstractindextext

\IEEEpeerreviewmaketitle

\ifCLASSOPTIONcompsoc
\IEEEraisesectionheading{\section{Introduction}\label{sec:introduction}}
\else
\section{Introduction}
\label{sec:introduction}
\fi

For an agent or system suffering from disturbances, its control input, as a defender, desires to complete a specified control mission by determining control policy to reject the influences of antagonistic input, i.e., non-cooperative disturbances that intend to disrupt the mission. This is known as zero-sum games or min-max problems \cite{Hi,OC}. In real-world applications, agent dynamics may be unknown. In order to make such an agent perform in the target trajectories exhibited by a target agent with optimal policy, optimal control theory assumes that the performance cost function is known, and RL \cite{RLbook,FADP} based optimal tracking control methods \cite{adg1,Moff,BH} compute optimal policy by observing states and control actions without knowing the system dynamics, where a standard iterative form for RL is known as policy iteration (PI)
\cite{RLbook,FADP,MTC,PI}. However, in real interactions, operators may not know the appropriate specified cost functions, i.e., the weights on states and inputs. As a result, these optimal control methods may not obtain the expected control performance or even be used.

Instead of manually selecting cost function weights, many efforts have been made on constructing cost function weights.
Inverse optimal control (IOC) and inverse RL (IRL) construct cost function weights given system control behaviors. Sometimes they are referred to as the same thing \cite{AIOC,r64,new2}, but they may differ in structure and how they are applied \cite{hv}.

Assuming a stable control system, IOC constructs a cost function concerning which the system behavior is optimal. The cost function is constructed in the framework of Lyapunov stability condition for continuous-time (CT) systems \cite{stIOC0,IOC,stIOC1,stIOC3} and discrete-time (DT) systems \cite{DTI,DTI1,r62} where \cite{r62} considers finite horizon. 
Online IOC methods to determine cost function in the infinite and finite horizon are studied in \cite{r63,r61}. IOC is also used to verify the effectiveness of the proposed control laws in \cite{stIOC2,stIOC4}. These works do not consider min-max or zero-sum games, but \cite{ZSDTIOC} does. They all require system dynamics, which cannot be applied directly to systems with unknown dynamics.

IRL generally reconstructs reward and cost functions from expert demonstrations of the optimal policy. It is usually applied to apprenticeship learning and imitation learning problems of Markov decision processes (MDPs) \cite{ComIRL,ComIRL3,ComIRL1,meirl,new2} where a learner seeks to imitate the demonstrations by learning the unknown expert's reward function from the observed demonstrations. IRL methods construct reward function since reward function is a more succinct, robust, and transferable definition for the task than the policy mapping from states to actions. Lyapunov stability is not necessarily considered here.

IRL has also been developed for trajectory tracking and imitation problems of differential systems in \cite{TCIRL,r65}, where \cite{TCIRL} uses a bilevel structure  (also see \cite{ComIRL,ComIRL3,nIOC}). That is, an optimal control problem is solved repeatedly in the inner loop. This two-loop iteration is computationally expensive. All of these works are model-based and do not consider min-max or zero-sum games. The work \cite{r66} makes an effort for data-driven control by estimating model parameters before adopting the model-based IRL method. Unlike them, without the need for model identification, our previous studies \cite{qian1,qian2} propose completely model-free IRL methods that use merely system data, but not for zero-sum games. Our work \cite{bosen} considers zero-sum games but propose an IRL method using a two-loop iteration structure and partial system dynamics.

This paper considers an expert-learner zero-sum game, that is, a learner agent suffering from non-cooperative disturbances with unknown dynamics expects to mimic the behaviors of the expert agent of optimal policy. As the solution, we propose a new interaction called zero-sum game IRL, namely a novel data-driven off-policy IRL algorithm for expert-learner zero-sum games of differential systems. It consists of a game solution correction modified from the standard RL and a cost function weight reconstruction using the standard IOC. Using only the behavior data of the expert and learner, a learner agent learns the unknown cost function objective and the optimal control policy to mimic the expert's behavior. This algorithm does not need to know or identify system models and performs a single-loop learning procedure without solving optimal problems repeatedly in inner loops. Moreover, no initial stabilizing policy is needed to start the iteration. The properties and effectiveness of the proposed data-driven are well-analyzed.

\noindent \textbf{Notations}. $\|{\cdot}\|$ is the Euclidean norm. $I_n$ is $n\times n$ identity matrix, and diag\{a,b,..\} is diagonal matrix with $a,b,..$ in diagonal line. For a vector $x=[x_1, \ \cdots, \ x_n]^T \in \mathbb{R}^{n}$, $\hat x\triangleq [x_{1}^2$, $2x_{1}x_{2}$, $\cdots$, $2x_{1}x_{n}$, $x_{2}^2$, $\cdots$, $2x_{n-1}x_{n}$, $x_{n}^2]^T$. For a matrix $A=\{a_{ij}\}\in \mathbb{R}^{m\times n}$, $\text{vec}(A)\triangleq [a_{11}$, $a_{12}$, $\cdots$, $a_{1n}$, $a_{21}$, $\cdots$, ${a_{m,n-1}}$, $a_{mn}]^T$.

\section{IRL Problem Formulation}\label{Form}

We consider two dynamical agents. A target expert agent exhibits the demonstrations optimally associated with an expert performance cost function. A learner agent attempts to determine the unknown cost function objective of the expert agent and mimic its behavior. The learner agent only knows the target agent's control actions and state behavior but does not know its performance cost function and system dynamics.

\subsection{Target optimal control}

Consider a target expert agent
\begin{align}
\dot x_T=Ax_T+Bu_T+Dd_T,
\label{e2}
\end{align}
where $x_T \in {\mathbb{R}^n}$ is the target state, $u_T \in {\mathbb{R}^m}$ is the target input and $d_T \in {\mathbb{R}^z}$ is the non-cooperative disturbance. Matrices $A$, $B$, and $D$ have appropriate dimensions. The pair $(A, B)$ is assumed to be controllable. 

According to \cite{OC}, the input of target \eqref{e2} is $u_T=-K_Tx_T$ that minimizes the following target performance cost function against $d_T$
\begin{align}
V_T(x_T)=\int^{\infty}_t(x_T^TQ_Tx_T+u_T^TR_Tu_T-\gamma_T^2d_T^Td_T)d\tau, 
\label{e3}
\end{align}
where $Q_T=Q_T^T>0$ and $R_T=R_T^T>0$ are weights, $\gamma_T>0$ is attenuation factor. The input $u_T$ and the worst disturbance $d_T^*$ are given by 
\begin{subequations}
\begin{align}
\begin{split}
u_T=-K_Tx_T=-R_T^{-1}B^TP_Tx_T,
\label{e5}
\end{split}\\
\begin{split}
d_T^*=L_Tx_T=\frac{1}{\gamma_T^2}D^TP_Tx_T,
\label{e6}
\end{split}
\end{align}
\end{subequations}
and form the Nash equilibrium
\begin{align}
V_T(x_T)^*=\min\limits_{u_T}\max\limits_{d_T}\int^{\infty}_t r_T d\tau=x_T^TP_Tx_T
\label{e4}
\end{align}
where $r_T=x_T^TQ_Tx_T+u_T^TR_Tu_T-\gamma_T^2d_T^Td_T$, and $P_T=P_T^T>0$ satisfies the target Bellman equation
\begin{align}
&(Ax_T+Bu_T+Dd_T)^TP_Tx_T+x_T^TP_T(Ax_T+Bu_T+Dd_T)\nonumber\\
&+x_T^TQ_Tx_T+u_T^TR_Tu_T-\gamma_T^2d_T^Td_T=0,
\label{e4b}
\end{align}
and the target game algebraic Riccati equation (GARE)
\begin{align}
A^TP_T+P_TA+Q_T-P_TBR_T^{-1}B^TP_T+\frac{1}{\gamma_T^2}P_TDD^TP_T=0.
\label{e7}
\end{align}

The GARE guarantees the uniqueness of stabilizing optimal control policy \eqref{e5} given the performance cost function \eqref{e3}.

\subsection{Learner dynamics and IRL control problem}

Consider a learner agent to be controlled
\begin{align}
\dot x=Ax+Bu+Dd,
\label{e1}
\end{align}
where $x \in {\mathbb{R}^n}$, $u \in {\mathbb{R}^m}$, $d\in {\mathbb{R}^z}$ are the learner's state, control input and disturbance, respectively.

\begin{assumption} The target cost function $V_T$ \eqref{e3} and the target control policy in \eqref{e5} are unknown to the learner \eqref{e1}. That is, the weights $Q_T$, $R_T$, $\gamma_T$ in $V_T$, the optimal strategy $K_T$ in \eqref{e5}, the worst disturbance $d_T^*$, $L_T$ and GARE solution $P_T$ are all unknown.
\end{assumption}

\begin{assumption} The learner knows the target behaviour data $x_T$, $u_T$ and $d_T$.
\end{assumption}

\begin{definition} 
\textbf{Expert-Learner Zero-Sum Game.} By using the behaviour data of the expert \eqref{e2} and the learner itself \eqref{e1}, the learner desires to reconstruct the unknown performance cost function \eqref{e3} to exhibit the same control actions $u_T$ in \eqref{e5} and states $x_T$ as the expert \eqref{e2}.\hfill $\Box$
\end{definition} 

The learner \eqref{e1} will be stabilized and perform the same way as the target \eqref{e2} does if $K_T$ in \eqref{e5} is applied to the learner \eqref{e1} with bounded $d_T=d$. Hence, our control goal is to determine the unknown cost function objective \eqref{e3} to produce the optimal control input $u^*=-K^*x$ with $K^*=K_T$ using only target data of $x_T, u_T, d_T$ and learner data of $x, u, d$.

\section{Model-Based IRL Framework}

In this section, we develop a novel model-based IRL framework for learner \eqref{e1} to determine the cost function \eqref{e3} and use this knowledge to compute control input $u(t)$, such that its behavior trajectories of $u(t),x(t)$ mimic the observed target trajectories of $u_T(t),x_T(t)$. In Section 5, we will finally propose a data-driven IRL algorithm that does not need any system dynamics.

\subsection{Learner optimal control}

Let us take a review of the zero-sum game \cite{OC} of the learner \eqref{e1} with an arbitrary given cost function
\begin{align}
V(x)=\int^{\infty}_t(x^TQx+u^TRu-\gamma^2d^Td)d\tau,
\label{e8}   
\end{align}
where $Q=Q^T>0$, $R=R^T>0$, and $\gamma>0$. The optimal input $u_o$ and the worst $d_w$ are given by
\begin{align}
\begin{split}
u_o=-Kx=-R^{-1}B^TPx,\label{e10}\\
\end{split}\\
\begin{split}
d_w=Lx=\frac{1}{\gamma^2}D^TPx,\label{e11}
\end{split}
\end{align}
and $(u_o, d_w)$ forms the Nash equilibrium 
\begin{align}
V^*(x)=\min\limits_{u}\max\limits_{d} \int^{\infty}_t r d\tau=x^TPx,
\label{e9a}   
\end{align}
where $r=x^TQx+u^TRu-\gamma^2d^Td$, and $P=P^T>0$ satisfies the following learner GARE
\begin{align}
A^TP+PA+Q-PBR^{-1}B^TP+\frac{1}{\gamma^2}PDD^TP=0.
\label{e12} 
\end{align}

\subsection{Expert-learner zero-sum game solution}\label{Opc}

We now present a theorem to show the conditions that the solution to the expert-learner zero-sum game must satisfy.

\begin{theorem}\label{th1}
If the cost weights $Q, R, \gamma$ and the solution $P$ that satisfy the GARE \eqref{e12} also satisfy the following equation
\begin{align}
(A-BK_T)^TP+&P(A-BK_T)\nonumber\\+&Q+\frac{1}{\gamma^2}PDD^TP+K_T^TRK_T=0,
\label{e16}   
\end{align}
then, the corresponding control strategy $K$ in \eqref{e10} equals the target strategy $K_T$ in \eqref{e5}.
\end{theorem}
\noindent \textbf{Proof:} Rewrite \eqref{e12} with $K_T$ in \eqref{e5} and $K$ in \eqref{e10} as
\begin{align}
&(A-BK_T)^TP+P(A-BK_T)+Q+\frac{1}{\gamma^2}PDD^TP\nonumber\\
&+K_T^TRK+K^TRK_T-K^TRK=0.
\label{e14}   
\end{align}
By subtracting \eqref{e14} from \eqref{e16}, we have
\begin{align}
&K_T^TRK_T-K_T^TRK-K^TRK_T+K^TRK\nonumber\\
&=(K_T-K)^TR(K_T-K)=0.
\label{e15}    
\end{align}
Since $R>0$, \eqref{e15} concludes that $K=K_T$. \hfill\ensuremath{\Box}

\subsection{Learning rules for cost function and game control policy}

To find the $Q,R,\gamma,P$ satisfying Theorem \ref{th1}, we select $\gamma>0$ and $R>0$ and propose an iterative procedure based on Theorem \ref{th1} so as to learn the weight $Q$, the game control policy $P$, and consequently $K$ in \eqref{e10} and $L$ in \eqref{e11}.

First, we apply a modified PI to correct $P$ using \eqref{e16}. Set current iteration step as $i, i=0,1,\cdots,$ and give the estimates $Q^i$ and $L^i$. Then the iterative form of \eqref{e16}, i.e., \eqref{a11} in Algorithm 1 is presented below to obtain $P^i$. Then optimal control (\eqref{e10} and \eqref{e11}) is used to update the strategy $K^{i+1}$ and $L^{i+1}$ based on the corrected $P^i$ by \eqref{a12} and \eqref{a13}, respectively.

Now we must update the cost function weight estimate $Q^{i+1}$ based on the corrected $P^i$. By IOC \cite{IOC}, taking the iterative form of GARE \eqref{e12} yields \eqref{a14} in Algorithm 1 presented as follows.

\begin{algorithm}[h] 
	\caption{Model-based IRL algorithm for expert-learner zero-sum games.} 
	
	\begin{enumerate}[{\textbf{Step}} 1:]
	
		\item Initialize with $R>0$ and $\gamma>0$, $Q^0>0$, and $L^0=0$, and set $i=0$.

		\item \textbf{(Game policy correction)} Update policy $P^i$ by
        \begin{flalign}
        &(A-BK_T)^TP^{i}+P^{i}(A-BK_T)&\nonumber\\
        &=-Q^i-K_T^TRK_T-\gamma^2(L^{i})^TL^{i}.
        \label{a11}
        \end{flalign}

        \item \textbf{(Input update)} Update control input and disturbance gain based on \eqref{e10} and \eqref{e11}
        \begin{subequations}
        \begin{flalign}
        u^{i+1}&=-K^{i+1}x=-R^{-1}B^TP^{i}x,&\label{a12}\\
        d^{i+1}&=L^{i+1}x=\frac{1}{\gamma^2}D^TP^{i}x.
        \label{a13}
        \end{flalign}
        \end{subequations}

        \item \textbf{(Cost function weight construction)} Update $Q^{i+1}$ by    
        \begin{flalign}
        Q^{i+1}=&- A^TP^{i}-P^{i}A&\nonumber\\
        &+ (K^{i+1})^TRK^{i+1}-\gamma^2(L^{i+1})^TL^{i+1}.&
        \label{a14}
        \end{flalign}

        \item Stop if it converges. Otherwise, set $i=i+1$ and repeat steps 2 to 4.
		
	\end{enumerate}
	\label{al1}
\end{algorithm}

\begin{remark} In Algorithm 1, step 4 is  the standard IOC computation based on Lyapunov stability condition. Note also that if we set $Q^i=Q_T$ and $K_T=K^i$ in \eqref{a11}, then steps 2-3 are the standard RL PI. As such, Algorithm \ref{al1} combines modified PI with IOC to solve the IRL problem. 
\end{remark}

\section{Analysis of Algorithm \ref{al1}}\label{proof}

The convergence, stability, and optimality of the proposed Algorithm \ref{al1} are analyzed here. It is also shown that Algorithm \ref{al1} may not converge to a unique solution $(P^i, Q^i)$ even if all solutions give the correct target strategy $K^{i}=K_T$.

\subsection{Convergence analysis}

\begin{theorem}\label{th2} 
i). With an initial $Q^0$ such that $0<Q^0\leq \hat Q$ and $L^0=0$ where $\hat Q>0$ is a solution to Theorem \ref{th1} associated with the $R>0$ and $\gamma>0$, Algorithm \ref{al1} converges. ii). As $i\rightarrow \infty$, the solutions $Q^{i}, P^{i}, K^{i}, L^{i}$ converge to $Q^*, P^*, K^*, L^*$ that satisfy
\begin{subequations}
\begin{flalign}
&\hspace{-2mm}A^TP^*+P^*A+Q^*\!-\!P^*BR^{-1}B^TP^*\!+\!\frac{1}{\gamma^2}P^*DD^TP^*=0,\label{e19}\\  
&\hspace{-2mm}K^*=K_T=R^{-1}B^TP^*,\label{e19a}\\
&\hspace{-2mm}L^*=\frac{1}{\gamma^2}D^TP^*.\label{e19b}    
\end{flalign} 
\end{subequations}
iii). The solutions $Q^*, P^*, K^*, L^*$ satisfy Theorem 1.
\end{theorem}
\noindent \textbf{Proof:} \textbf{i). Convergence proof.}
First, we prove that Algorithm 1 solves an increasing sequence $Q^i$ for all $i=0,1,\cdots$. Substituting \eqref{a12} for $K^i$ into \eqref{a14} for $Q^i$ yields
\begin{align}
A^TP^{i-1}+P^{i-1}A=(K^i)^TRK^i-Q^i-\gamma^2(L^i)^TL^i,
\label{t10}
\end{align}
which can be rewritten as
\begin{align}
&(A-BK_T)^TP^{i-1}+P^{i-1}(A-BK_T)\nonumber\\
&=(K^i)^TRK^i-Q^i-\gamma^2(L^i)^TL^i-(K^i)^TRK_T-K_T^TRK^i.
\label{t11}
\end{align}
Subtracting \eqref{a11} from \eqref{t11} gives
\begin{align}
&(A-BK_T)^T(P^{i-1}-P^{i})+(P^{i-1}-P^{i})(A-BK_T)\nonumber\\
&=(K^i-K_T)^TR(K^i-K_T).
\label{t12}
\end{align}
It follows from $(K^i-K_T)^TR(K^i-K_T)\geq 0$ and Hurwitz $A-BK_T$ that $P^{i-1}\leq P^{i}$ holds for all iterations. Each pair of $(P^i, Q^{i+1})$ satisfies \eqref{a14}, and they uniquely correspond to each other. This is followed by the known fact that $P^{i-1}\leq P^i$ if $Q^{i}\leq Q^{i+1}$ \cite{ARE}. Therefore, $Q^{i+1}\geq Q^{i}>0$ holds for $i=0,1,\cdots$, and $Q^{i+1}= Q^{i}$ holds if and only if $K_T=K^{i+1}$. Note that achieving $K_T=K^{i+1}$ is the goal of the algorithm.

Now we show that $Q^i$ is bounded by an upper bound. Let $\hat Q>0$ and $\hat P>0$ be a group of solution to Theorem \ref{th1}. That is
\begin{subequations}
\begin{align}
\begin{split}
A^T\hat P+\hat PA+\hat Q-\hat PBR^{-1}B^T\hat P+\frac{1}{\gamma^2}\hat PDD^T\hat P=0,\label{t17}\\
\end{split}&\\
\begin{split}
K_T=R^{-1}B^T\hat P,\ \hat L=\frac{1}{\gamma^2}D^T\hat P.
\label{t18a}\\
\end{split}
\end{align}
\end{subequations}
Rewriting \eqref{t17} using \eqref{t18a} yields
\begin{align}
(A-BK_T)^T\hat P+\hat P(A-BK_T)+\hat Q+K_T^TRK_T+\gamma^2\hat L^T\hat L=0.
\label{t113}
\end{align}
If $Q^i+\gamma^2(L^{i})^T L^i\leq \hat Q + \gamma^2\hat L^T \hat L$ holds, then \eqref{a11} and \eqref{t113} will solve $0<P^i\leq \hat P$ with Hurwitz $A-BK_T$. With \eqref{e5}, \eqref{a12} and \eqref{t18a}, AREs \eqref{a14} and \eqref{t17} can be rewritten as
\begin{subequations}
\begin{align}
&\hspace{-1mm}(A-BK^{i+1})^T P^i+ P^i(A-BK^{i+1})&\nonumber\\
&\hspace{-1mm}=-Q^{i+1}-(K^{i+1})^TRK^{i+1}-\gamma^2(L^{i+1})^T(L^{i+1}), &\label{t15}\\
&\hspace{-1mm}(A-BK^{i+1})^T\hat P+\hat P(A-BK^{i+1})\nonumber\\
&\hspace{-1mm}=K_T^TRK_T-\hat Q-\gamma^2\hat L^T\hat L-(K^{i+1})^TRK_T-K_T^TRK^{i+1},
\label{t16}
\end{align}
\end{subequations}
respectively. Since \eqref{a14} ensures Hurwitz $A-BK^{i+1}$, subtracting \eqref{t15} from \eqref{t16} and using $0<P^i\leq \hat P$ obtains
\begin{align}
&(A-BK^{i+1})^T(\hat P-P^{i})+(\hat P-P^{i})(A-BK^{i+1})\nonumber\\
&=(Q^{i+1}+\gamma^2(L^{i+1})^TL^{i+1})-(\hat Q+\gamma^2\hat L^T\hat L)\nonumber\\
&\quad +(K^{i+1}-K_T)^TR(K^{i+1}-K_T)\leq 0.
\label{t19}
\end{align}
Therefore, $Q^{i+1}+\gamma^2(L^{i+1})^TL^{i+1}\leq \hat Q+\gamma^2\hat L^T\hat L$ holds.

By deduction, it is inferred that initializing Algorithm 1 with a $Q^0$ such that $0<Q^0\leq \hat Q$ and $L^0=0$, then $Q^i>0, i=0,1,\cdots$ will be increasing with an upper bound. Therefore, Algorithm 1 converges.

\noindent\textbf{ii). Converged solutions satisfy \eqref{e19}, \eqref{e19a}, \eqref{e19b}.}

Substituting \eqref{a14} into \eqref{a11} yields
\begin{align}
&A^TP^{i}+P^{i}A-P^iBR^{-1}B^TP^i\nonumber\\
&=A^TP^{i+1}+P^{i+1}A-P^{i+1}BK_T\!-\!K_T^TB^TP^{i+1}+K_T^TRK_T.
\label{t13}
\end{align}
Taking $P^{i+1}=P^i=P^*$ as converged value, \eqref{t13} becomes
\begin{align}
(K_T-K^*)R(K_T-K^*)=0,
\label{t14}
\end{align}
where $K^*=R^{-1}B^TP^*$. Since $R>0$, \eqref{t14} implies $K_T=K^*$, which is exactly \eqref{e19a}. The converged $P^*$ produces the converged $L^*$ using \eqref{a13} as shown in \eqref{e19b} and the converged $Q^*$ using \eqref{a14} as shown in \eqref{e19}.

\noindent\textbf{iii). Converged solutions satisfy Theorem 1.}

Rewriting \eqref{e19} with \eqref{e19a} yields 
\begin{align}
&(A-BK_T)^TP^*+P^*(A-BK_T)+Q^*\nonumber\\
&+K_T^TRK_T+\frac{1}{\gamma^2}P^*DD^TP^*=0,
\label{t110}   
\end{align}
which is exactly \eqref{e16}. Obviously, \eqref{e19} is exactly \eqref{e12}. Therefore, $Q^*, P^*, K^*$ satisfy Theorem \ref{th1}.  \hfill\ensuremath{\Box}

\subsection{Stability and optimality analysis}

We now prove the stability of Algorithm \ref{al1} in Theorem \ref{th3}, and optimality and Nash equilibrium in Theorem \ref{th6} 

\begin{theorem}\label{th3}
Each iteration of Algorithm \ref{al1} exponentially stabilizes the learner agent \eqref{e1} with $d=0$.
\end{theorem}
\noindent\textbf{Proof:} Rewriting \eqref{a11} with \eqref{e5} and \eqref{a12} yields
\begin{align}
A^TP^i+P^iA+\tilde Q^i-P^iBR^{-1}B^TP^i+\gamma^2(L^i)^TL^i=0.
\label{t33}
\end{align}
where $\tilde Q^i=Q^i+(K^{i+1}-K_T)^TR(K^{i+1}-K_T)$. With $(K^{i+1}-K_T)^TR(K^{i+1}-K_T)\geq0$ and $Q^i>0$, then $\tilde Q^i>0$. It is obvious that $P^i$ solved by \eqref{t33} or equivalently \eqref{a11} is a symmetrical positive definite matrix satisfying
\begin{align}
A^TP^i+P^iA-P^iBR^{-1}B^TP^i+\gamma^2(L^i)^TL^i<0,
\label{t34}
\end{align}
and one has
\begin{align}
\dot V^i(x,P^i)&=x^T(A-BK^{i+1})^TP^ix+x^TP^i(A-BK^{i+1})x\nonumber\\
&<-x^TP^iBR^{-1}B^TP^ix-\gamma^2x^T(L^i)^TL^ix<0.
\label{t35}
\end{align}
That is, $P^i$ yields stabilizing $K^{i+1}$ by \eqref{a12} for learner \eqref{e1} with $d=0$.

Considering \eqref{a12} and \eqref{t34}, $Q^{i+1}$ in \eqref{a14} satisfies
\begin{align}\nonumber
&Q^{i+1}+\gamma^2(L^{i+1})^TL^{i+1}\\
&=-(A^TP^i+P^iA-P^iBR^{-1}B^TP^i)>0.
\label{t37}
\end{align}
Using $Q^{i+1}+\gamma^2(L^{i+1})^TL^{i+1}>0$ in \eqref{a11} would still make \eqref{t35} hold for the next iteration. Therefore, provided $Q^0>0$ \eqref{t35} will hold for all $i=0,1,\cdots$.
\hfill\ensuremath{\Box}

Before optimality analysis of Algorithm 1, we now give a lemma of importance which extends the idea of classic IOC \cite{IOC} to two-player zero-sum games.

\begin{lemma} Consider the two-player learner agent \eqref{e1} with $x(t_0)=x_0,\ t\geq t_0$ and \eqref{e8} and \eqref{e9a}. Assume there exists a positive definite symmetric matrix $P\in {\mathbb{R}^{n\times n}}$ such that
\begin{align}
A^TP+PA-PBR^{-1}B^TP+\frac{1}{\gamma^2}PDD^TP<0.
\label{l11}
\end{align}
Then, with the optimal feedback control input $u_o$ and the worst disturbance $d_w$ such that
\begin{align}
u_o=-R^{-1}B^TPx,\ d_w=\frac{1}{\gamma^2}D^TPx,
\label{l12}
\end{align}
and the cost function weight 
\begin{align}
Q=-(A^TP+PA-PBR^{-1}B^TP+\frac{1}{\gamma^2}PDD^TP),
\label{l14}
\end{align}
the saddle point $(u_o,d_w)$ makes the cost value function \eqref{e8} reach the Nash equilibrium
\begin{align}
V(x_0,u_o,d)\leq V(x_0,u_o,d_w)\leq V(x_0,u,d_w).
\label{l15a}
\end{align}
\end{lemma}
\noindent\textbf{Proof:} First, $V(x)$ in \eqref{e8} can be represented with $P>0$ as
\begin{align}
V(x)=x^TPx\geq 0, \ V(0)=0.
\label{l16}
\end{align}
It follows from \eqref{l12} and \eqref{l14} that $H(x,u_o,d_w)=0$ \cite{OC} where the Hamiltonian is
\begin{align}
H(x,u,d)=&x^TQx+u^TRu-\gamma^2d^Td+(Ax+Bu+Dd)^TPx\nonumber\\
&+x^TP(Ax+Bu+Dd).
\label{l17}
\end{align}
One writes
\begin{align}
H(x,u,d)&=H(x,u,d)-H(x,u_o,d_w)\nonumber\\
&=(u-u_o)R(u-u_o)-\gamma^2(d-d_w)^T(d-d_w),
\label{l18}
\end{align}
and hence 
\begin{align}
H(x_0,u_o,d)\leq H(x_0,u_o,d_w)\leq H(x_0,u,d_w).
\label{l19}
\end{align}
Based on \cite{OC}, we obtain the conclusion \eqref{l15a}. \hfill\ensuremath{\Box}

\begin{theorem}\label{th6}
The converged solutions $Q^*,P^*,K^*,L^*$ obtained by Algorithm \ref{al1} shown in Theorem \ref{th2} yield Nash equilibrium of the value function $V(x)$ in \eqref{e8} such that
\begin{align}
V(x_0,u^*,d)\leq V(x_0,u^*,d^*)\leq V(x_0,u,d^*),
\label{t61}
\end{align}
where $u^*=-K^*x$ and $d^*=L^*x$.
\end{theorem}
\noindent\textbf{Proof:} 
It follows from Theorem \ref{th1} that $Q^i>0$ holds for all $i=0,1,\cdots$. Thus one has the converged $Q^*>0$ and 
\begin{align}
A^TP^*+P^*A-P^*BR^{-1}B^TP^*+\frac{1}{\gamma^2}(P^*)^TDD^TP^*<0,
\label{t63}
\end{align}
from \eqref{e19}, which means that the converged $P^*$ satisfies \eqref{l11} in Lemma 1. Also, the converged control strategy $K^*$ in \eqref{e19a} and disturbance gain $L^*$ in \eqref{e19b} satisfy \eqref{l12} in Lemma 1. This indicates that \eqref{l19} also holds for $u^*$ and $d^*$, namely the Nash equilibrium \eqref{t61} holds. \hfill\ensuremath{\Box}

\subsection{Non-uniqueness of solution}

In fact, the $Q^*, R, \gamma, P^*$ satisfying \eqref{e19}-\eqref{e19b} that explain the same strategy $K^*=K_T$ may not be unique and can be different from the actual target values $Q_T, R_T, \gamma_T, P_T$ shown in \eqref{e6} and \eqref{e7}. This multi-solution phenomenon is known as the ill-posedness property, which is well-analyzed for DT ARE in \cite{ill1} and coupled ARE in \cite{ill2}. In the next result, we characterize the relationship between $Q_T, R_T, \gamma_T, P_T$, and $Q^*, R, \gamma, P^*$ for CT GARE and show the conditions for the occurrence of this phenomenon.

\begin{theorem}\label{th4n} 
Recall $Q_T$, $R_T$, $\gamma_T$, $P_T$ satisfying \eqref{e7} and \eqref{e6}, and let $Q_o$, $R_o$, $P_o$ satisfy
\begin{subequations}
\begin{align}
&B^TP_o=R_oR_T^{-1}B^TP_T,\label{t41n}\\
&Q_o+A^TP_o+P_oA-K_T^TR_oK_T+\frac{1}{\gamma_T^2}P_TDD^TP_T-\frac{1}{\gamma^2}P^*DD^TP^*\nonumber\\
&=0,
\label{t42n}
\end{align}
\end{subequations}
where $R_o=R_T-R$. Then any $Q^*=Q_T-Q_o$ and $P^*=P_T-P_o$ satisfy \eqref{e19}-\eqref{e19b}. 
\end{theorem}
\noindent\textbf{Proof:} 
Subtracting \eqref{t42n} from \eqref{e7} and using \eqref{e6} yields
\begin{align}
A^T(P_T-P_o)&+(P_T-P_o)A+(Q_T-Q_o)\nonumber\\
&-K_T^T(R_T-R_o)K_T+\frac{1}{\gamma^2}P^*DD^TP^*=0.
\label{t44n}
\end{align}
Using $P^*=P_T-P_o$, $R_o=R_T-R$, and $R>0$ in \eqref{t41n} gives
\begin{align}
K^*=R^{-1}B^TP^*=R_T^{-1}B^TP_T=K_T,\label{t45n}
\end{align}
which is \eqref{e19a}. Substituting it into \eqref{t44n} yields \eqref{e19} and \eqref{e19b}.

This proves the relationship between the obtained solution $Q^*, R, \gamma, P^*$ and the expert's $Q_T, R_T, \gamma_T, P_T$. We observe that $P_o$, $R_o$, $Q_o$ satisfying \eqref{t41n} and \eqref{t42n} can be nonzero. That is, $Q^*, R, \gamma, P^*$ associate optimally with the same strategy as $Q_T, R_T, \gamma_T, P_T$, i.e., $K_T=K^*$, but $Q_T\neq Q^*$, $R_T\neq R$, $\gamma_T\neq \gamma$. Therefore, there could be multiple solutions to \eqref{e19} to generate a $K^*$ in \eqref{e19a} equal to the target $K_T$ in \eqref{e5}. \hfill\ensuremath{\Box}

The following corollary shows a special case of the $Q^*, R, \gamma$ of Theorem 5 which gives $V(x)^*=cV_T(x_T)^*$.

\begin{corollary}\label{co1} 
With scalar $c>0$, any $Q^*=cQ_T$, $R=cR_T$, $\gamma=\sqrt{c}\gamma_T$ would yield $V(x)^*=cV_T(x_T)^*$ in \eqref{e4} and \eqref{e9a}, and they optimally associate with the same $K^*$  as the expert such that $K^*=K_T$ in \eqref{e19a} and \eqref{e5}.
\end{corollary}
\noindent\textbf{Proof:} Bring such $Q^*, R, \gamma$ into \eqref{e19}, since $Q_T, R_T, \gamma_T$ satisfy \eqref{e7}, then one has $P^*=cP_T$ and $K^*=K_T$. Then, using this result in \eqref{e9a} for $V(x)^*$ and comparing it with $V_T(x_T)^*$ in \eqref{e4} shows that $V(x)^*=cV_T(x_T)^*$. \hfill\ensuremath{\Box}

\section{Data-Driven Off-Policy IRL Algorithm}\label{V}

Algorithm \ref{al1} relies on the system dynamics $A, B, D$ and the target strategy $K_T$. To remove this requirement, we develop here a data-driven IRL algorithm for expert-learner zero-sum games based on Algorithm \ref{al1}, which only requires the data $x_T, u_T, d_T$ of the target agent \eqref{e2} and $x, u, d$ of the learner agent \eqref{e1}. To accomplish this, we use two techniques similar to integral RL \cite{Moff,MTC} and off-policy RL \cite{Moff,JY}. The end result is Algorithm \ref{al2}.

\subsection{Data-driven game policy correction}\label{ct1}

In order to update $P^i$, $K^{i+1}$ and $L^{i+1}$ in \eqref{a11}-\eqref{a13} using only target data $x_T, u_T, d_T$, inspired by the idea of off-policy integral RL technique \cite{Moff,JY}, rewrite \eqref{e2} as
\begin{align}
\dot x_T=Ax_T-BK^ix_T+Dd_T+B(u_T+K^ix_T).
\label{ed1}
\end{align}
Using \eqref{ed1} and \eqref{a11} one writes
\begin{align}
&\dot x_T^TP^{i}x_T+x_T^TP^{i}\dot x_T\nonumber\\
&=(Ax_T-BK^ix_T)^TP^{i}x_T+x_T^TP^{i}(Ax_T-BK^ix_T)\nonumber\\
&\quad+2(u_T+K^ix_T)^TB^TP^{i}x_T+2d_T^TD^TP^{i}x_T\nonumber\\
&=-x_T^TQ^ix_T-x_T^TK_T^TRK_Tx_T-\gamma^2x_T^T(L^{i})^TL^{i}x_T\nonumber\\
& \quad +2x_T^TK_T^TB^TP^ix_T+2u_T^TB^TP^{i}x_T+2d_T^TD^TP^{i}x_T.
\label{ed2}
\end{align}
Using \eqref{a12}, \eqref{a13}, \eqref{e5} in \eqref{ed2} and integrating both sides from $t$ to $t+T$, where $T>0$ is the integral time period, obtains \eqref{a21} in Algorithm 2 to be presented, by which $P^i$, $K^{i+1}$ and $L^{i+1}$ are updated simultaneously. Similar to \cite{Moff}, probing noise $e$ is added to $u_{T}$, i.e., $u_{T}=-K_Tx_T+e$, for the persistence of excitation condition in merely learning process. It is not needed anymore when solutions converge.
Unlike \eqref{a11}-\eqref{a13}, \eqref{a21} does not need the knowledge of agent dynamics $A,B,D$ or the strategy $K_T$ in \eqref{e5}.

\subsection{Data-driven cost function weight reconstruction}

In order to update $Q^{i+1}$ in \eqref{a14} using only data $x,u,d$, inspired by the integral RL technique \cite{MTC}, multiplying both sides of \eqref{a14} by $x$ and adding and subtracting terms $u^TB^TP^ix$ and $d^TD^TP^ix$, \eqref{a14} can be rewritten as
\begin{align}
x^TQ^{i+1}x=&- [(Ax+Bu+Dd)^TP^{i}x+x^TP^{i}(Ax+Bu+Dd)\nonumber\\
&- x^T(K^{i+1})^TRK^{i+1}x+\gamma^2x^T(L^{i+1})^TL^{i+1}x\nonumber\\
&-2u^TB^TP^ix-2d^TD^TP^ix],
\label{ed6}
\end{align}
where $u$ can be generated by any stabilizing policy and $d$ can be random and different from $d_T$ in the learning process. Substituting \eqref{e1}, \eqref{a12} and \eqref{a13} into \eqref{ed6} and integrating it gives \eqref{a22} in Algorithm 2 below.
Using $P^i$, $K^{i+1}$ and $L^{i+1}$ obtained by \eqref{a21}, \eqref{a22} equivalently replaces \eqref{a14} in Algorithm \ref{al1} to calculate $Q^{i+1}$ without knowing any system dynamics.

\begin{algorithm}[h] 
	\caption{Data-driven off-policy IRL algorithm for expert-learner zero-sum games.} 

	\begin{enumerate}[{\textbf{Step}} 1:]

		\item Initialize with $R>0$, $\gamma>0$, $Q^0>0$, and $L^0=0$, and collect system data generated by any stabilizing control input $u$. Set $i=0$.

		\item \textbf{(Game policy correction)} Update policy $P^i$, control strategy $K^{i+1}$ and disturbance gain $L^{i+1}$ by
        \begin{flalign}
       & x_T(t+T)^TP^{i}x_T(t+T)-x_T(t)^TP^{i}x_T(t)&\nonumber\\
        &	 -2\int_t^{t+T}e^TRK^{i+1}x_Td\tau-2\gamma^2\int_t^{t+T}d_T^TL^{i+1}x_T d\tau\nonumber\\
        &	 =-\int_t^{t+T}\big(x_T^TQ^ix_T+(u_T-e)^TR(u_T-e)\nonumber\\
        &\quad +\gamma^2x_T^T(L^{i})^TL^{i}x_T\big) d\tau.
        \label{a21}
        \end{flalign}

        \item \textbf{(Cost function weight construction)} Update cost function weight $Q^{i+1}$ by
        \begin{flalign}
       & \hspace{-2mm}  \int_t^{t+T}x^TQ^{i+1}x d\tau&\nonumber\\
        & \hspace{-2mm}   =- [x(t+T)^TP^{i}x(t+T)-x(t)^TP^{i}x(t)&\nonumber\\
        & \hspace{-2mm}   \quad -\int_t^{t+T}(2u^TRK^{i+1}x+x^T(K^{i+1})^TRK^{i+1}x) d\tau&\nonumber\\
        & \hspace{-2mm}   \quad - \! \gamma^2 \int_t^{t+T} \! (2d^TL^{i+1}x-x^T(L^{i+1})^TL^{i+1}x) d\tau].
        \label{a22}
        \end{flalign}

        \item Stop if it converges. Otherwise, set $i=i+1$ and repeat steps 2 to 4, .
    
	\end{enumerate}
	\label{al2}
\end{algorithm}

\begin{remark} Algorithm \ref{al2} does not need system dynamics. Moreover, it iterates in single loop indicated by $i$, no inner-loop iteration is needed.
\end{remark}

\subsection{Implementation and Analysis of Algorithm \ref{al2}}

In order to show how to implement data-driven IRL Algorithm \ref{al2} using only data, first, consider Kronecker product $a^TWb=(b^T\otimes a^T)\text{vec}(W)$ for \eqref{a21} and define the following operators, 
\begin{align}
&O_{x_Tx_T}=[x_{T1}^2,2x_{T1}x_{T2},...,x_{T2}^2,2x_{T2}x_{T3},...,x_{Tn}^2]^T;&\nonumber\\
&d_{x_Tx_T}=[O_{x_Tx_T}(t+T)-O_{x_Tx_T}(t),...,\nonumber\\
&\quad \quad \quad \ \ O_{x_Tx_T}(t+lT)-O_{x_Tx_T}(t+(l-1)T)]^T;\nonumber\\
&I_{x_Tx_T}=[\int_t^{t+T}O_{x_Tx_T} d\tau,...,\int_{t+(l-1)T}^{t+lT}O_{x_Tx_T} d\tau]^T;\nonumber\\
&I_{x_Tu_T}=[\int_t^{t+T}x_T \otimes u_T d\tau,...,\int_{t+(l-1)T}^{t+lT}x_T\otimes u_T d\tau]^T;\nonumber\\
&I_{x_Td_T}=[\int_t^{t+T}x_T\otimes d_T d\tau,...,\int_{t+(l-1)T}^{t+lT}x_T\otimes d_T d\tau]^T;\nonumber\\
&I_{x_Te}=[\int_t^{t+T}x_T \otimes e d\tau,...,\int_{t+(l-1)T}^{t+lT}x_T\otimes e d\tau]^T;\nonumber\\
&\Phi_p=[d_{x_Tx_T},-2I_{x_Te}(I_n\otimes R),-2\gamma^2I_{x_Td_T}]^T;\nonumber\\
&r^i=x_T^TQ^ix_T+(u_T-e)^TR(u_T-e)+\gamma^2x_T^T(L^{i})^TL^{i}x_T;\nonumber\\
&\Psi_p^i=-[\int_{t}^{t+T} r^i d\tau,...,\int_{t+(l-1)T}^{t+lT}r^id\tau]^T;\nonumber\\
&\hat P^i=[P^i_{11},P^i_{12},...,P^i_{22},P^i_{23},...,P^i_{nn}]^T,
\label{e30}
\end{align}
where $l$ is the group number of collected data and should be $l\geq \frac{n(n+1)}{2}+nm+nz$. Using batch least squares method \cite{MTC, JY, BLS}, $\hat P^i$, $K^{i+1}$, $L^{i+1}$ can be calculated by 
\begin{align}
\hspace{-2mm}\left[\!\!
{\begin{array}{*{20}{c}}
	{(\hat P^{i})^T},
	{\text{vec}(K^{i+1})^T},
	{\text{vec}(L^{i+1})^T}
	\end{array}} \right]^T\!\!\!=\!(\Phi_p^T\Phi_p)^{-1}\Phi_p^T\Psi_p^i.
\label{e31}
\end{align}

Similarly, for \eqref{a22}, we define
\begin{align}
&O_{xx}=[x_{1}^2,2x_{1}x_{2},...,x_{2}^2,2x_{2}x_{3},...,x_{n}^2]^T;\nonumber\\
&d_{xx}=[O_{xx}(t+T)-O_{xx}(t),...,\nonumber\\
&\quad \quad \ \ O_{xx}(t+kT)-O_{xx}(t+(k-1)T)]^T;\nonumber\\
&\hat q^i=[Q^i_{11},Q^i_{12},...,Q^i_{22},Q^i_{23},...,Q^i_{nn}]^T;&\nonumber\\
&\Phi_q=I_{xx}=[\int_t^{t+T}O_{xx} d\tau,...,\int_{t+(k-1)T}^{t+kT}O_{xx} d\tau]^T;&\nonumber\\
&I_q^{i+1}(t)=\int_t^{t+T}(2u^TRK^{i+1}x+x^T(K^{i+1})^TRK^{i+1}x) d\tau&\nonumber\\
        &\quad \quad \quad \quad + \gamma^2\int_t^{t+T}(2d^TL^{i+1}x-x^T(L^{i+1})^TL^{i+1}x) d\tau\nonumber\\
&\Psi_q^{i+1}=-d_{xx}\hat P^i+[I_q^{i+1}(t),...,I_q^{i+1}(t+(k-1)T)]^T,
\label{e33}
\end{align}
where $k$ is the group number of collected data and should be $k\geq \frac{n(n+1)}{2}$. Then, $Q^{i+1}$ can be uniquely solved by
\begin{align}
\hat q^{i+1}= (\Phi_q^T\Phi_q)^{-1}\Phi_q^T\Psi_q^{i+1}.
\label{e35}
\end{align}
By using \eqref{e31} and \eqref{e35}, we solve $P^i$, $Q^{i+1}$, $K^{i+1}$, $L^{i+1}$ in a data-driven mode.

Each step of Algorithm \ref{al1} yields a unique solution. Algorithm \ref{al2} is developed based on Algorithm \ref{al1}. To illustrate that the solution obtained by Algorithm \ref{al2} estimates the solution obtained by Algorithm \ref{al1}, we show that equations \eqref{e31} and \eqref{e35} yield unique solutions in the next result.

\begin{theorem}
If there exist $l_o>0$, $k_o>0$, for all $l\geq l_o$ and $k\geq k_o$, 
\begin{subequations}
\begin{align}
&\text{rank}([I_{x_Tx_T},I_{x_Tu_T},I_{x_Td_T}])=\frac{n(n+1)}{2}+nm+nz,&\label{t51}\\
&\text{rank}(I_{xx})=\frac{n(n+1)}{2},
\label{t52}
\end{align}
\end{subequations}
then, \eqref{e31} and \eqref{e35} solve unique solution, respectively.
\end{theorem}
\noindent\textbf{Proof:} First, we show that \eqref{e31} solves unique solution. This is aiming to show that 
\begin{align}
\Phi_p\Omega=0\label{t53}
\end{align}
has only the trivial solution $\Omega=0$. Now, we prove $\Omega=0$ by contradiction. Assume $\Omega=[X_{v}^T, Y_{v}^T, Z_{v}^T]^T \in {\mathbb{R}^{\frac{n(n+1)}{2}+nm+nz}}$ is a nonzero solution of \eqref{t53}, where $X_{v}\in {\mathbb{R}^{\frac{n(n+1)}{2}}}$, $Y_{v}\in {\mathbb{R}^{mn}}$, $Z_{v}\in {\mathbb{R}^{nz}}$. Then, $X_{v}, Y_{v}, Z_{v}$ uniquely determine matrices $X, Y, Z$ by $X_{v}=\hat X$, $Y_{v}=\text{vec}(Y)$ and $Z_{v}=\text{vec}(Z)$, respectively, where $X$ is a symmetrical matrix. 

Define
\begin{align}
In_x=[\int_t^{t+T}x_T\otimes x_T d\tau,...,\int_{t+(k-1)T}^{t+kT}x_T\otimes x_T d\tau]^T.
\label{t54}
\end{align}
Integrating \eqref{ed2} from $t$ to $t+T$ gives
\begin{align}
\Phi_p\Omega=In_{x}\text{vec}(E)+2 I_{x_Tu_T}\text{vec}(G)+2I_{x_Td_T}\text{vec}(F)\hspace{-0.5mm}=0,
\label{t55}
\end{align}
where 
\begin{subequations}
\begin{align}
&E=A^TX+XA-K_T^TRY-Y^TRK_T,\label{t56}\\
&G=B^TX-RY,\label{t57}\\
&F=D^TX-\gamma^2Z.
\label{t58}
\end{align}
\end{subequations}
Since $E$ is a symmetrical matrix, one has $In_{x}\text{vec}(E)=I_{x_Tx_T}\hat E$. Using this in \eqref{t55} yields
\begin{align}
\Phi_p\Omega=[I_{x_Tx_T},2I_{x_Tu_T},2I_{x_Td_T}]\left[
{\begin{array}{*{20}{c}}
	{\hat E}\\
	{\text{vec}(G)}\\
	{\text{vec}(F)}
	\end{array}} \right]=0.
\label{t510}
\end{align}
Under \eqref{t51}, we know that $[I_{xx},I_{xu},I_{xd}]$ has full column rank, and thus \eqref{t510} has only the solution $\hat E=0$, $\text{vec}(G)=0$ and $\text{vec}(F)=0$. That is,
\begin{subequations}
\begin{align}
&A^TX+XA-K_T^TRY-Y^TRK_T=0,\label{t511}\\
&B^TX=RY,\label{t512}\\
&D^TX=\gamma^2Z.
\label{t513}
\end{align}
\end{subequations}
Since $A-BK_T$ is Hurwitz, substituting \eqref{t512} and \eqref{t513} into \eqref{t511} gives $X=0$. This implies that $Y=0$ and $Z=0$ due to $R>0$ and $\gamma^2>0$. In summary, we have $\Omega=0$. However, this conflicts with the assumption that $\Omega$ is nonzero. Therefore, it concludes that under \eqref{t51}, \eqref{e35} solves unique solution. 

Second, from the integral RL work \cite{MTC}, we conclude that when \eqref{t52} holds for \eqref{a22}, $Q^{i+1}$ in \eqref{a22} can be uniquely determined by \eqref{e35} with collected data. \hfill\ensuremath{\Box}

\section{Simulation}

We show three simulation experiments, a first one of the data-driven Algorithm \ref{al2} to show its performance, a comparison simulation with the bilevel IRL method in \cite{bosen} to show the reduction of iteration steps of Algorithm \ref{al2}, and a second comparison simulation with the RL-based optimal tracking control method in \cite{Moff} to show the improvement of control performance with the cost function weights correction of Algorithm 2. 

\subsection{Simulation result of Algorithm \ref{al2}}\label{m1}

The system dynamics information of the target \eqref{e2} and the learner \eqref{e1} for simulation is
\begin{align}
A=\left[
{\begin{array}{*{20}{c}}
	{-1}& {2}\\
	{2.2}& {1.7}
	\end{array}} \right], B=\left[
{\begin{array}{*{20}{c}}
	{0}\\
	{3}
	\end{array}} \right], D=\left[
{\begin{array}{*{20}{c}}
	{1}\\
	{0}
	\end{array}} \right].
\label{e80}
\end{align}
For the target agent \eqref{e2}, the actual target cost function objective \eqref{e3} consists of $Q_T=\text{diag}\{8, 12\},
R_T=2I_1, \gamma_T=3$. The disturbance is $d_T=0.003\text{rand}(1)$. The expert's $L_T$, $K_T$ and $P_T$ are
\begin{align}
&K_T=[1.9869,  3.5779], L_T=[0.4162,  0.1472], \nonumber\\
&P_T=\left[
{\begin{array}{*{20}{c}}
	{3.7459 } &   {1.3246}\\
	{1.3246}  &  {2.3853}         
\end{array}}\right].
\label{e82}
\end{align}

For the learner, the behaviour strategy to generate data is $K_b=[1.2129, \ 2.2812]$, and the disturbance is $d=0.003\text{rand}(1)$. To start Algorithm 2, the initial weights for cost function, initial disturbance gain $L^0$, and integral time period $T$ are given by
\begin{align}
&Q^0=\text{diag}\{1, 0.5\},
R=I_1, \gamma=40, L^0=[0,\ 0], T=0.008s.
\label{e83}
\end{align}

\begin{figure}[htp]
	\centering
\includegraphics[width=0.5\textwidth]{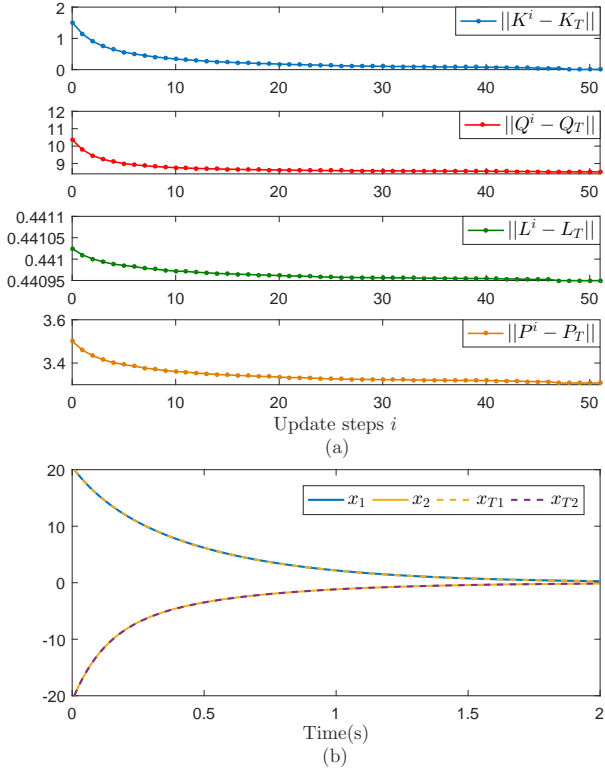}
	\caption{Convergence and imitation performance using Algorithm \ref{al2}}
	\label{dQPKL}
\end{figure}

The Fig. \ref{dQPKL}(a) captures the iteration process from the initial spot to the spot on $\|K^{i+1}-K_T\|\leq 0.01$. The final values of $K^i$, $Q^i$, $L^i$, and $P^i$ are $K^*$, $Q^*$, $L^*$, and $P^*$, respectively, as follows
\begin{align}
&K^*=\left[{\begin{array}{*{20}{c}}
{1.9827}  &  {3.5839}
\end{array}}\right],\nonumber\\
&Q^*=\left[
{\begin{array}{*{20}{c}}
	{2.2796} &   {2.6670}\\
	{2.6670}  &  {6.0151}      
\end{array}}\right], \nonumber\\
&L^*=10^{-3}\times\left[{\begin{array}{*{20}{c}}
{0.4021}  &  {0.4131}
\end{array}}\right], \nonumber\\
&P^*=\left[
{\begin{array}{*{20}{c}}
	{0.6441 } &   {0.6622}\\
	{0.6622}  &  {1.1968}         
\end{array}}\right],
\label{e86}
\end{align}
where $K^*$ closely approximates the target $K_T$ in \eqref{e82} with $\|K^*-K_T\|=0.0073$, while $Q^*$, $L^*$ and $P^*$ are not equal to $Q_T, L_T$ and  $P_T$ in \eqref{e82}, respectively. This is the multiple-solution phenomenon. In Fig. \ref{dQPKL}(b), the learner's state $x$ can mimic the trajectories of the target $x_T$ very well under the learned $K^*$. Therefore, the proposed Algorithm 2 can learn an appropriate cost function and optimal policy for the learner to mimic the target trajectories.

\subsection{Comparison simulation case 1}\label{m3}

This subsection shows the simulation results of the bilevel IRL method in \cite{bosen} that iterates in two-loop to show the reduction of computational complexity in terms of iteration steps. The same expert-learner system, initial parameters for Algorithm 2 in \eqref{e80}-\eqref{e83} are used for this comparison method 1.

\begin{figure}
	\centering	\includegraphics[width=0.5\textwidth]{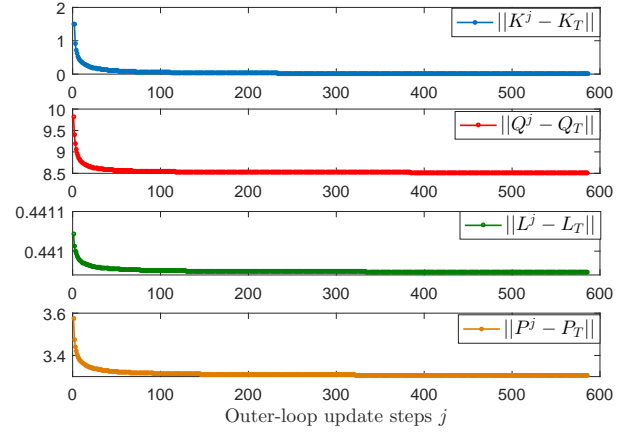}
	\caption{Convergence of $K^{j}, Q^{j}, L^{j}$, and $P^{j}$ using the comparison method 1}
	\label{duibi}
\end{figure}

Fig. \ref{duibi} also captures the iteration process from the initial spot to the spot on $\|K^{i+1}-K_T\|\leq 0.01$ as Fig. \ref{dQPKL} to show the difference in iteration steps of the two methods. The inner-loop iteration figures are omitted since they are too many to put here. In Fig. \ref{duibi}, the final values of $K^{j}, Q^{j}, L^{j}$, and $P^{j}$ of the outer-loop iterations are
\begin{align}
&K^*=\left[{\begin{array}{*{20}{c}}
{1.9822}  &  {3.5691}    
\end{array}}\right],\nonumber\\
&Q^*=\left[
{\begin{array}{*{20}{c}}
	{2.3186} &   {2.6974}\\
	{2.6974}  &  {6.0506}    
\end{array}}\right], \nonumber\\
&L^*=10^{-3}\times\left[{\begin{array}{*{20}{c}}
{0.4053}  &  {0.4130}    
\end{array}}\right], \nonumber\\
&P^*=\left[
{\begin{array}{*{20}{c}}
	{0.6486} &   {0.6607}\\
	{0.6607}  &  {1.1897}    
\end{array}}\right].    
\end{align}
where $K^*$ approximates the target $K_T$ in \eqref{e82} with $\|K^*-K_T\|=0.01$. Table \ref{tab1} shows that the total iteration steps of the method is 3370, including 587 outer-loop updates (See Fig. \ref{duibi}) and 2783 inner-loop updates, while Algorithm 2 iterates 51 steps in total (See Fig. \ref{dQPKL}(a)). The time of the learning process of Algorithm 2 is 4.08s, while that of the comparison method 1 is 169.936s. It is proportional to the amount of utilized collected data. Algorithm 2 uses 510 groups of data, and comparison method 1 uses 21242 groups of data. Therefore, Algorithm 2 costs much fewer data and time than the bilevel comparison method 1.

\begin{table}[htp]
	\caption{Iteration steps and learning time of Algorithm 2 and the comparison method 1}
	\label{tab1}
	\centering
	\begin{tabular}{cccc}
		\hline\hline
	   Methods	&  Total updates & Learning time \\	\hline
		Algorithm 2 & 51 & 4.08s \\ 
		Comparison method 1 & 3370 & 169.936s \\
		\hline\hline
	\end{tabular}
\end{table}

\subsection{Comparison simulation case 2}\label{m4}

In this subsection, the typical RL-based optimal tracking control method \cite{Moff} for linear disturbed systems, which computes optimal control policy given cost function weights, is simulated to show the advantage of Algorithm 2 in control performance by computing both optimal control policy and cost function weights. 

\begin{figure}[htp]
	\centering	\includegraphics[width=0.5\textwidth]{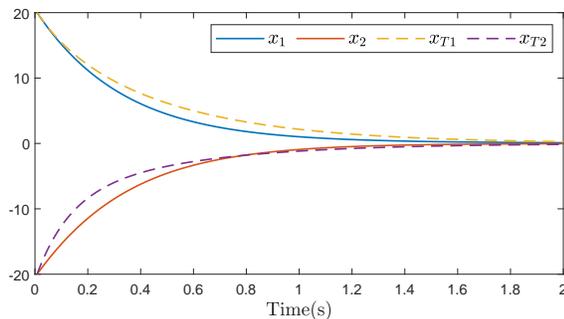}
	\caption{Imitation performance of learner's $x$ to target $x_T$ using the comparison method 2}
	\label{duibirl}
\end{figure}

The same system and cost weights shown in \eqref{e80}-\eqref{e83} and discount factor $\alpha=0.9$ are used. The obtained optimal control law is $u^*=-[ 1.1760  \  1.9139  \  1.0044  \  1.7639][x^T,r^T]^T$, and the corresponding imitation performance in Fig. \ref{duibirl} is not as good as that of Algorithm 2 in Fig. \ref{dQPKL}(b). By evenly sampling the trajectory data, the imitation performance of the two methods is quantified by the error index defined as follows
\begin{align}
Te= \frac{1}{n}\sum\limits_{i = 1}^n {\sqrt {\frac{1}{{a }}\sum\limits_{k = {1}}^{ a} {|{x_i}(kT) - {x_{Ti}}(kT){|^2}} } }\nonumber
\end{align}
where $n=2$, $a=250$, and $T=0.008s$. As shown in Table \ref{tab2}, $Te$ of Algorithm 2 is much smaller than that of the comparison method 2. The reason is that Algorithm 2 can correct the given cost function weights when it is inappropriate, but the comparison method 2 cannot. Algorithm 2 thus obtains much better imitation performance. 

\begin{table}[htp]
	\caption{Imitation index of Algorithm 2 and the comparison method 2}
	\label{tab2}
	\centering
	\begin{tabular}{cccc}
		\hline\hline
	   Methods	&  Error index $Te$  \\	\hline
		Algorithm 2 & 0.0162  \\ 
		Comparison method 2 & 1.4461 \\
		\hline\hline
	\end{tabular}
\end{table}

\section{Conclusion} 

This paper proposes a novel data-driven off-policy IRL approach to determine both cost function and optimal control policy to stabilize a learner agent suffering from non-cooperative disturbances by mimicking a target agent's trajectories using data of both agents. The proposed approach does not need any system dynamics and guarantees stability, Nash optimality, and imitation performance with single-loop iteration. The rigorous theoretical proofs and simulation experiments verify its effectiveness.

\ifCLASSOPTIONcaptionsoff
  \newpage
\fi



\bibliographystyle{IEEEtran}
\bibliography{IEEEabrv,bibfile}

\end{document}